\documentclass[runningheads]{llncs}
\usepackage[T1]{fontenc}
\usepackage{graphicx}
\usepackage{orcidlink}
\begin{document}
%
%\title{LLM-as-a-Fuzzy-Judge: Aligning an LLM-Powered Medical Education System with Human Preferences}
\title{LLM-as-a-Fuzzy-Judge: Fine-Tuning Large Language Models as a Clinical Evaluation Judge with Fuzzy Logic}
\titlerunning{LLM-as-a-Fuzzy-Judge}
\author{Weibing Zheng\orcidlink{0000-0002-6496-1906}  \and Laurah Turner\orcidlink{0000-0002-4567-1313} 
\and Jess Kropczynski\orcidlink{0000-0002-7458-6003} \and Murat Ozer 
\and Tri Nguyen \and
Shane Halse\orcidlink{0000-0002-0799-6715}
%\and 
% XXXXXX\inst{2}\orcidID{0000-0002-4567-1313}\and 
% XXXXXX\inst{2}\orcidID{0000-0002-4567-1313}\and 
% XXXXXX\inst{2}\orcidID{0000-0002-4567-1313}
 }
\authorrunning{Zheng, W et al.}
\institute{University of Cincinnati, Cincinnati, OH, USA  
% \and {Department of Medical Education, and Biostatistics, Health Informatics and Data Sciences, University of Cincinnati College of Medicine, Cincinnati, Ohio, USA }
\email{zhengwb@mail.uc.edu}
}
\maketitle              % typeset the header of the contribution
%
%\begin{abstract}
% The abstract should briefly summarize the contents of the paper in

\begin{abstract}
Clinical communication skills are critical in medical education, and practicing and assessing clinical communication skills on a scale is challenging. Although LLM-powered clinical scenario simulations have shown promise in enhancing medical students' clinical practice, providing automated and scalable clinical evaluation that follows nuanced physician judgment is difficult. This paper combines fuzzy logic and Large Language Model (LLM) and proposes LLM-as-a-Fuzzy-Judge to address the challenge of aligning the automated evaluation of medical students' clinical skills with subjective physicians' preferences. LLM-as-a-Fuzzy-Judge is an approach that LLM is fine-tuned to evaluate medical students' utterances within student-AI patient conversation scripts based on human annotations from four fuzzy sets, including Professionalism, Medical Relevance, Ethical Behavior, and Contextual Distraction. The methodology of this paper started from data collection from the LLM-powered medical education system, data annotation based on multidimensional fuzzy sets, followed by prompt engineering and the supervised fine-tuning (SFT) of the pre-trained LLMs using these human annotations. The results show that the LLM-as-a-Fuzzy-Judge achieves over 80\% accuracy, with major criteria items over 90\%, effectively leveraging fuzzy logic and LLM as a solution to deliver interpretable, human-aligned assessment. This work suggests the viability of leveraging fuzzy logic and LLM to align with human preferences, advances automated evaluation in medical education, and supports more robust assessment and judgment practices. The GitHub repository of this work is available at \url{https://github.com/2sigmaEdTech/LLMAsAJudge}.

\end{abstract}

\section{Introduction}

Large language models (LLMs) have rapidly transformed the landscape of medical education, allowing new forms of virtual patient simulation, automated feedback, and clinical reasoning assessment~\cite{nori2023capabilities}~\cite{gilson2023does}. The LLMs-powered clinical training system (2-Sigma) has been pivoted into medical education at the College of Medicine, University of Cincinnati. The LLM-powered pivoted medical education system has generated thousands of text conversation scripts. As LLM-powered medical systems are increasingly integrated into different educational scenarios, the need for robust and human-aligned clinical evaluation frameworks has become critical. Traditional clinical evaluation methods, often based on binary (Pass/Fail) or discrete scoring, struggle to capture the nuanced, context-dependent judgment required in real-world clinical training~\cite{wang2023chatgpt}.

A key challenge in the deployment of LLM-powered systems is to ensure that their outputs align with human values, ethical standards, and domain-specific expectations. For example, most medical students follow the rules to perform as professional physicians when practicing their clinical skills; however, some interactions with patients are not appropriate and unprofessional. Human expert evaluation remains the gold standard, but it is resource-intensive and subject to variability~\cite{zhang2023language}. Moreover, the complexity of clinical interactions often involves ambiguous or borderline cases that do not fit neatly into true/false or correct/incorrect categories.

Fuzzy logic offers a powerful framework for modeling the uncertainty and gradation inherent in human judgment~\cite{herrera2000fusion}. By representing evaluation criteria as fuzzy sets, it becomes possible to capture the spectrum of possible judgment, such as professionalism, medical relevance, ethical behavior, and contextual distraction, more faithfully than with rigid categorical labels~\cite{phuong2001fuzzy}~\cite{suzukiFuzzyLogicSystems2024}. 

\begin{figure}
\centering
\includegraphics[width=0.63\textwidth]{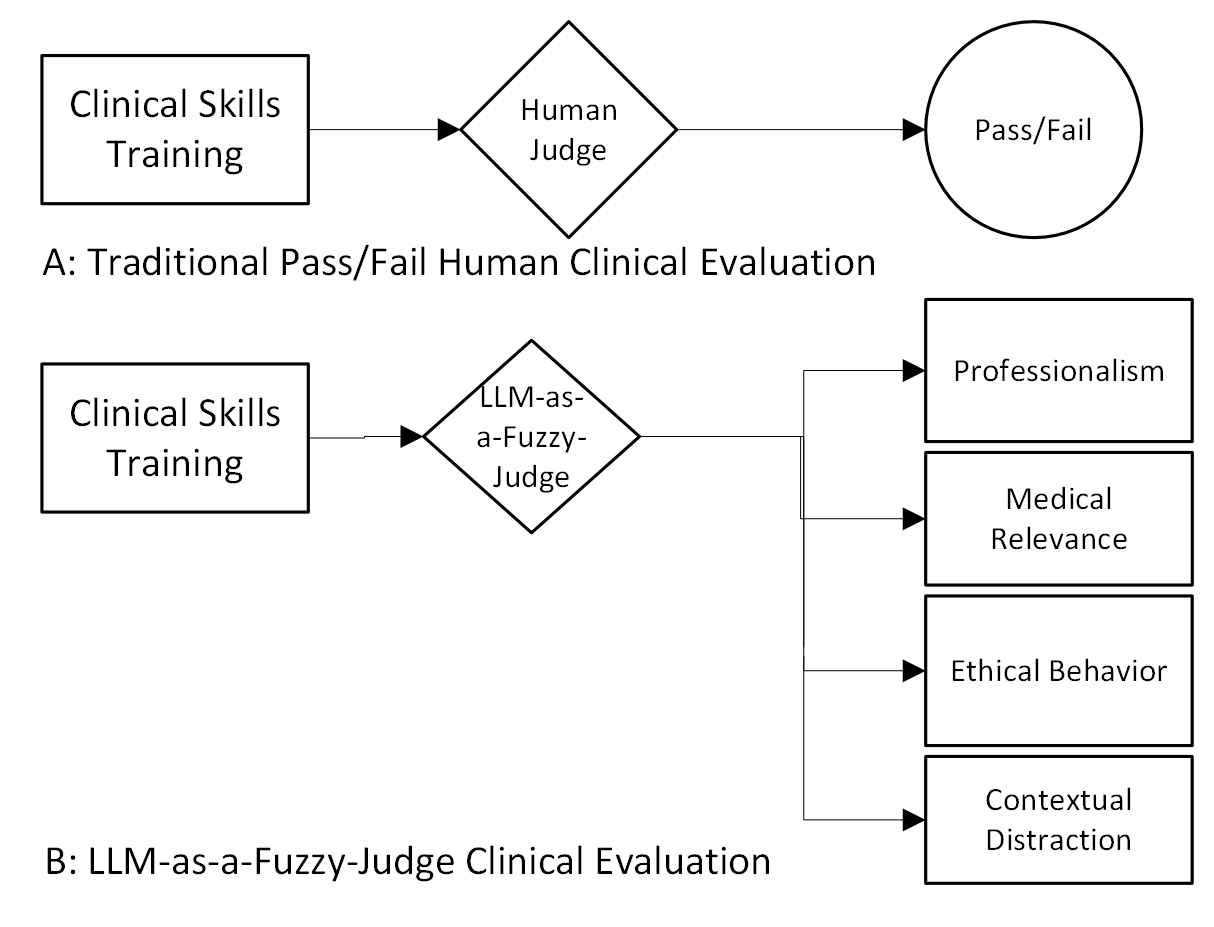}
\caption{A. Traditional Evaluation vs B. LLM-as-a-Fuzzy-Judge Evaluation}\label{fig1:fuzzyjudge}
\end{figure}

In this study, we introduce \textbf{LLM-as-a-Fuzzy-Judge}, shown as Figure~\ref{fig1:fuzzyjudge}. This hybrid evaluation framework combines supervised fine-tuning (SFT) and prompt engineering to align the LLM judgment with the judgment of human experts in medical education. Our approach leverages a large, multi-annotator dataset of medical student-AI patient conversations, annotated across multiple fuzzy criteria. We demonstrate the hybrid model achieves strong agreement with human judges, provides nuanced and interpretable feedback, and advances the state-of-the-art in automated, human-aligned evaluation for medical education.

\section{Background and Related Work}

\subsection{Large Language Models}
LLMs are special Generative Artificial Intelligence (GenAI) algorithms in the field of Natural Language Processing (NLP) and primarily focus on text and language. LLMs ~\cite{changSurveyEvaluationLarge2024} are built on neural networks on enormous amounts of text data (corpus) and can generate text content. LLMs are models of intelligent generative systems that can understand, manipulate, and process human language and generate human-like text to mimic human communication. The training process of LLMs is to help models learn the mathematical probability from the statistical patterns in the huge and vast sequential text, and to predict the next highest possible word in the sequence. In 2017, the transformer architecture proposed in ``Attention is All You Need"~\cite{vaswaniAttentionAllYou2017} marked a turning point, leading to more LLMs emerging, such as Bidirectional Encoder Representations (BERT) and Generative Pre-trained Transformer (GPT). The transformer architecture~\cite{vaswaniAttentionAllYou2017} is at the heart of the most advanced LLMs. The encoder-decoder NLP structure processes sequential input and output words with self-attention mechanisms. It assigns different words with different weights based on their importance in a sentence and allows the model to generate relevant and coherent text based on the calculated probability. LLMs know a lot from their trained text; however, if the data from the training text on the query and responses do not exist or are old, LLMs can cause hallucination problems, which means the responses are not accurate, not related, or not right~\cite{naveedComprehensiveOverviewLarge2024}~\cite{kasneciChatGPTGoodOpportunities2023}. It is a major challenge for the production-level use of LLMs. To help mitigate the issue, a downstream LLM task is to fine-tune the pre-trained LLMs from the upstream model checkpoint on a specific task, such as text classification, text generation, and text summarization. The fine-tuning process is to train the model on a specific task with a smaller dataset and adjust the weights of the pre-trained model to make it more suitable for the specific task. The fine-tuning process is usually faster than training a model from scratch and can achieve better performance on specific tasks. Prompts are the input queries that users enter into the chat box to interact with LLMs. Prompt engineering is about the engineering process of improving prompts to get a better and reliable output from the LLMs. Fine-tuning and prompt engineering processes are important steps in achieving a better performance of LLMs for a specific task~\cite{marvin2023prompt}~\cite{pornprasit2024fine}.

\subsection{LLMs in Medical Education and Simulation}
LLMs have recently been adopted in medical education for tasks such as virtual patient simulation, clinical reasoning training, and automated feedback generation. Studies have shown that LLMs can simulate patient responses and clinical scenarios, helping medical students develop diagnostic and communication skills~\cite{nori2023capabilities}~\cite{gilson2023does}. These models have shown promise in passing medical licensing exams and providing realistic, context-aware interactions, although concerns remain regarding factual accuracy and ethical considerations.

\subsection{Human Evaluation and Annotation in AI Systems}
Human evaluation remains a gold standard for assessing AI-related text, especially in high-stakes domains like healthcare. Annotation frameworks often involve expert evaluators who assess the text on multiple dimensions, such as relevance, safety, and professionalism~\cite{zhang2023language}. Recent work has highlighted the importance of multi-criteria and fuzzy (non-binary) annotation schemes to capture the nuanced judgments required in medical and educational contexts~\cite{wang2023chatgpt}~\cite{zhang2023language}.

\subsection{Fuzzy Logic and Fuzzy Set Theory in Assessment}
Fuzzy logic provides a mathematical framework for modeling uncertainty and gradation in human judgment, making it well suited for educational and clinical evaluation~\cite{zadeh1965fuzzy} ~\cite{ribeiro2014fif}. In medical education, fuzzy set theory has been used to evaluate student performance, clinical decision-making, and communication skills, allowing more nuanced feedback than binary or discrete scoring systems~\cite{phuong2001fuzzy}~\cite{suzukiFuzzyLogicSystems2024}. This approach aligns with the real-world complexity of clinical interactions, where judgments often fall on a spectrum.

\subsection{Aligning LLMs with Human Preferences}
Recent advances in aligning LLMs with human preferences have focused on techniques such as reinforcement learning from human feedback (RLHF), SFT with annotated data, and prompt engineering~\cite{ouyang2022training}. These methods aim to ensure that the LLM outputs reflect human values, ethical standards, and domain-specific requirements. In medical education, alignment is particularly critical to ensure safety, professionalism, and relevance in AI-generated feedback and simulation~\cite{wang2023chatgpt}~\cite{gilson2023does}.

\section{Methodology}

\subsection{An LLM-Powered Medical Education System: 2-Sigma}
The data for this study were collected from 2-Sigma, an LLM-powered precision medical education system. 2-Sigma is designed to improve medical students' clinical reasoning skills through simulated patient interactions to mimic real-world clinical encounters~\cite{zhengUserCenteredIterativeDesign2025}. In this system, medical students engage in conversations with LLM-simulated AI patients, navigating various clinical scenarios. The conversations are structured as multi-turn dialogues, where students assume the role of physicians and the LLM is configured as the patient. Medical students enter text to diagnose an AI patient to collect the patient's symptoms, medical history, utilize labs, tests, and interventions to get more patient's healthy data, and make decisions on the primary hypothesis and differential hypothesis. The system aims to improve students' clinical reasoning skills by providing a realistic and interactive learning environment.

\subsection{Data Collection}
All conversation logs are stored in structured SharePoint tables in 2-Sigma. The conversation table includes fields such as Conversation ID, Case, Jailbreak ID, Conversation Pair, User Message, and Assistant Message. The Conversation ID serves as a unique identifier for each conversation, while the Case field indicates the specific clinical scenario being simulated. The Jailbreak ID tracks attempts to bypass or manipulate AI safeguards, and the Conversation Pair numbers the sequence of interactions between the medical student and AI patient. The User Message field captures the student's questions or statements directed at the AI patient, while the Assistant Message field contains the AI patient's responses. 

We collected a dataset of 2,302 conversation logs from 2-Sigma between medical students and AI patients' interactions. From these conversation logs, individual medical student utterances were extracted as the primary data points for human judges to evaluate. The dataset was designed to cover a wide range of clinical scenarios, ensuring diversity and complexity in the interactions.

\subsection{Human Preferences and Fuzzy Annotations}

The utterance of each medical student was annotated by seven judges, including human medical professionals and experts. They evaluated student inputs based on four fuzzy criteria: \textbf{Professionalism}, \textbf{Medical Relevance}, \textbf{Ethical Behavior}, and \textbf{Contextual Distraction}. All fuzzy criteria were defined with a fuzzy set. These four criteria and fuzzy sets were discussed and defined by the research team, including medical education experts and AI researchers. Fuzzy sets were designed to capture the complexity and nuance of human judgment in clinical interactions. The fuzzy criteria and fuzzy sets were defined as follows.
\begin{itemize}
    \item \textbf{Professionalism} refers to maintaining appropriate conduct, language, and respect for boundaries, with violations including inappropriate comments, unprofessional tone, and breaches of confidentiality. This can include unethical behavior, such as actions that do not prioritize patient safety and consent, with red flags including harmful recommendations, neglect of symptoms, or any form of abuse. The fuzzy set for professionalism includes three levels: \textbf{1. Unprofessional, 2. Borderline, 3. Appropriate}.
    \item \textbf{Medical Relevance} tracks whether students remain focused on the case, identifying irrelevant questions, unnecessary exams, or misdiagnoses. The fuzzy set for medical relevance includes three levels: \textbf{1. Irrelevant, 2. Partially relevant, 3. Relevant}.
    \item \textbf{Ethical Behavior} ensures actions prioritize patient safety and consent, with red flags including harmful recommendations, neglect of symptoms, or any form of abuse. The ethical behavior contains five levels: \textbf{1. Dangerous, 2. Unsafe, 3. Questionable, 4. Mostly safe, 5. Safe}.
    \item \textbf{Contextual Distraction} assesses whether each message is relevant to the ongoing conversation and builds logically on previous exchanges between the student and the AI "patient." Violations occur when a message strays from the established context, introduces unrelated topics, or creates distractions that shift focus from the case discussion. This ensures that each interaction remains focused and contributes meaningfully to the case without sidestepping irrelevant or distracting content. Contextual distraction has four levels: \textbf{1. Highly distracting, 2. Moderately distracting, 3. Questionable, 4. Not distracting}.
\end{itemize}

We duplicated the data table into eight Excel sheets and sent it to eight expert judges. Each judge would annotate all four criteria based on the fuzzy sets. The judges were instructed to evaluate the student utterances independently, without discussing their evaluations with each other. This approach ensured that each judge's assessment was based solely on their expertise and understanding of the fuzzy criteria. The judges were also provided with a detailed description of the fuzzy sets and informed on how to apply them to the students' utterances. This was done to ensure consistency in the evaluation process and to help human judges understand the nuances of each fuzzy criterion. After data cleaning and formatting, we obtained a final dataset of 2,302 student utterances, each annotated by seven human judges in the four fuzzy criteria, with the condition that if they coded the same, then kept the same annotation; otherwise, picked the most frequent annotation. We follow the standard way to split the dataset into 70\% for training, 20\% for testing, and 10\% for validating. 

\subsection{LLM-as-a-Fuzzy-Judge Model Design}
Since the dataset is text-based and the task is to evaluate the student utterances, we choose a few text-specific pre-trained base models as the backbone for the LLM-as-a-Fuzzy-Judge model. The model is designed to predict fuzzy labels for each of the four criteria based on the student's utterance. The model architecture consists of an encoder-decoder structure, where the encoder processes the input text, and the decoder generates the fuzzy labels.

We propose a hybrid LLM-as-a-Fuzzy-Judge model that combines SFT and prompt engineering. The model is designed to leverage the strengths of both approaches, enabling it to learn from human annotations.
\begin{itemize}
    \item \textbf{Supervised Fine-Tuning (SFT)}: A pre-trained LLM is fine-tuned on the annotated dataset. Each input is paired with its corresponding fuzzy labels for all four criteria. The model is trained in a multi-task setting, simultaneously predicting the fuzzy level for each criterion, enabling it to learn complex evaluation patterns from human judges.
    \item \textbf{Prompt Engineering}: During inference, we use few-shot prompting by constructing input prompts that include several annotated examples. This guides the LLM to generate judgments consistent with human annotations, leveraging the model's in-context learning ability and improving generalization to new cases.
\end{itemize}
The hybrid approach aims to maximize both predictive accuracy and interpretability, allowing the model to adapt to evolving evaluation standards.

\subsection{Experimental Setup}
The annotated dataset is randomly divided into training, validation, and test sets. So, for the total of 2,303 rows of data, the training set has 1,611 rows, the validation set has 231 rows, and the test set contains 461 rows.  The fine-tuned model is evaluated in the test set using multiple metrics:
\begin{itemize}
    \item \textbf{Accuracy} and \textbf{weighted F1-score} for each fuzzy criterion
    \item \textbf{Agreement with human judges}, measured by sample confidence score
\end{itemize}
We compare the hybrid model's performance with two baselines: a pure prompt-based approach and a simple baseline (e.g., majority class prediction). Ablation studies are conducted to assess the individual contributions of fine-tuning and prompt engineering. Additionally, qualitative analysis is performed on challenging or ambiguous cases to illustrate the model's reasoning and alignment with human preferences.

This methodology ensures a robust and interpretable evaluation of the LLM-as-a-Fuzzy-Judge, supporting its application in the evaluation of medical education.

\section{LLM-as-a-Fuzzy-Judge Model Implementation}

\subsection{LLM Architecture and Fine-Tuning Process}
We utilize a state-of-the-art pre-trained LLM from Hugging Face, such as google-bert/bert-base-uncased, emilyalsentzer/Bio-ClinicalBERT, microsoft/deberta-v3-base, roberta-large, meta-llama/Meta-Llama-3.1-8B-Instruct, and mistralai/Mistral-7B-Instruct-v0.3 as the backbone for the Fuzzy Judge. The model is adapted for multi-task classification, with separate output heads for four fuzzy criterion. Each input sample consists of the medical student's message.

For SFT, the annotated dataset is formatted so that each input is paired with its corresponding fuzzy labels for all four criteria. The model is trained to minimize the sum of cross-entropy losses across all tasks, enabling it to learn complex, nuanced evaluation patterns from human judges. Hyperparameters such as learning rate, batch size, and epochs are optimized using the validation set.

\subsection{Prompt Engineering Strategies}
During inference, we take advantage of prompt engineering to further align the model output with human judgment. Few-shot prompting is used by constructing input prompts that include several annotated examples, each showing a student message and the corresponding fuzzy labels. This approach allows the LLM to generate judgments consistent with human annotations, even for cases not seen during training.

\textbf{Example prompt:}
\begin{quote}
% \texttt{

Follow the examples to rate Professionalism, Medical Relevance, Ethical Behavior, and Contextual Distraction.\\

Example 1: \\
Student: "Describe this dizziness - does it feel like the room was spinning or just like a fading to black
" \\
Professionalism: Appropriate \\
Medical Relevance: Relevant \\
Ethical Behavior: Safe \\
Contextual Distraction: Not distracting \\

Example 2: \\
Student: "im asking you that idiot" \\
Professionalism: Unprofessional \\
Medical Relevance: Irrelevant \\
Ethical Behavior: Mostly safe \\
Contextual Distraction: Moderately distracting \\

Now, evaluate the following: \\
Student: "Does the chest pain spread anywhere else?" \\

% }
\end{quote}
The LLM is expected to complete the fuzzy labels for the new input, following the pattern established by the examples.

\subsection{Training and Inference Pipeline}
The overall pipeline consists of the following steps:
\begin{enumerate}
    \item \textbf{Data Preprocessing:} Clean and format the annotated dataset, ensuring consistent representation of the conversation context and fuzzy labels. Clean some starting language, like " please start the conversation", and remove no used columns and only keep "User Message".
    \item \textbf{Supervised Fine-Tuning:} Train the LLM on the training set using multi-task objectives, validating performance on the validation set.
    \item \textbf{Prompt Construction:} For inference, dynamically construct prompts with a set of representative annotated examples from the training data.
    \item \textbf{Prediction:} For each new student input, use the fine-tuned LLM with the constructed prompt to predict fuzzy labels for all criteria.
    \item \textbf{Post-processing:} Aggregate predictions if needed (e.g., for ensemble or majority voting), and compare outputs with human annotations for evaluation.
\end{enumerate}

This implementation ensures that the LLM-as-a-Fuzzy-Judge leverages both explicit supervision and in-context learning, maximizing alignment with human evaluators in medical education scenarios. The detailed steps and code are in the GitHub repository \url{https://github.com/2sigmaEdTech/LLMAsAJudge}.

\section{Evaluation}

\subsection{Quantitative Results}
We evaluate the LLM-as-a-Fuzzy-Judge in the test set using multiple quantitative metrics for each fuzzy criterion: Professionalism, Medical Relevance, Ethical Behavior, and Contextual Distraction. The primary metrics include accuracy, weighted avg, weighted F1 score, and confidence to measure agreement with human judges. 

Table~\ref{tab:quantitative} summarizes the model's best performance across all criteria. The hybrid model (SFT + prompt engineering) is compared against two baselines: a pure prompt-based approach and a majority class predictor. The results demonstrate that the hybrid model consistently outperforms both baselines, achieving higher agreement with human annotations and better classification metrics across all fuzzy criteria.

\begin{table}[h]
\centering
\caption{Quantitative evaluation of LLM-as-a-Fuzzy-Judge on the test set}
\label{tab:quantitative}
\begin{tabular}{|l|c|c|c|c|}
\hline
\textbf{Criterion} & \textbf{Accuracy} & \textbf{Weighted avg} & \textbf{Weighted F1 Score} & \textbf{Best Model} \\
\hline
Professionalism & 0.8395 & 0.870 & 0.805 & Hybrid \\
Medical Relevance & 0.820 & 0.838 & 0.8207 & Hybrid \\
Ethical Behavior & 0.824 & 0.860 & 0.8403 & Hybrid \\
Contextual Distraction & 0.833 & 0.839 & 0.8321 & Hybrid \\
\hline
\end{tabular}
\end{table}

The study indicates that both SFT and prompt engineering contribute to performance gains, with the combination producing the highest agreement with human judges.

\subsection{Qualitative Results}
To further assess the alignment of the model with human judgment, we perform qualitative analysis on challenging and ambiguous cases from the test set. Selected case studies illustrate how the LLM-as-a-Fuzzy-Judge reasons about nuanced student inputs and produces fuzzy labels that closely match human annotations.

For example, in cases where student messages are borderline between "Appropriate" and "Unprofessional," the model often provides intermediate or context-aware judgments, reflecting the uncertainty present in real-world evaluation. Similarly, for ethical dilemmas or partially relevant questions, the model's outputs demonstrate an understanding of subtle distinctions, such as labeling a message as "Questionable" rather than strictly "Safe" or "Unsafe." For example, here are some sample distraction model predictions and high confidence. "Text: Chest xray | Pred: 3 | Confidence: 0.9917
Text: what if i told you there was nothing that could be done, because our cardiologists are stuck in Israel | Pred: 0 | Confidence: 0.9802."

These qualitative results highlight the model's ability to generalize beyond seen examples and to provide interpretable, human-aligned feedback in complex medical education scenarios.

Overall, the evaluation demonstrates that the LLM-as-a-Fuzzy-Judge achieves strong quantitative performance and produces nuanced, interpretable judgments that align well with expert human evaluators.

\section{Discussion}

The evaluation demonstrates that the LLM-as-a-Fuzzy-Judge achieves good alignment with human expert judgments across the four fuzzy criteria over 80\% accuracy, with the main criteria fuzzy level over 90\% accuracy. The hybrid approach, which combines SFT and prompt engineering, consistently outperforms both prompt-only and majority-class baselines in terms of accuracy, weighted F1 score, and weighted average. The model can capture nuanced, non-binary distinctions in professionalism, medical relevance, ethical behavior, and contextual distraction, reflecting the complexity of real-world clinical education.

A key strength of the hybrid approach is its ability to leverage both explicit human supervision and the in-context reasoning capabilities of LLMs. SFT enables the model to internalize expert annotation patterns, while prompt engineering allows for flexible adaptation to new or ambiguous cases. This combination results in improved generalization and interpretability.

However, the approach also has limitations. The quality and diversity of annotated data directly impact model performance, and the need for expert annotation can be resource-intensive. Also, some fuzzy criteria set levels are raw, and the data points are small. Different judges may have different opinions on annotations.  Additionally, while prompt engineering enhances flexibility, it may introduce variability in output depending on prompt construction. The performance of the model may also be constrained by the inherent biases and limitations of the underlying LLM. 

The LLM-as-a-Fuzzy-Judge provides a scalable and interpretable framework for evaluating medical student interactions with AI-simulated patients. By modeling LLMs as fuzzy judgments like human judges, the system can offer more nuanced feedback to learners and support faculty in assessment tasks. This approach also advances the broader field of AI evaluation by demonstrating how hybrid human-AI systems can align automated judgments with expert standards in high-stakes domains.

While this work focuses on medical education, the hybrid LLM-as-a-Fuzzy-Judge framework applies to other domains where nuanced, non-binary evaluation is essential, such as legal reasoning, education, customer service, and ethics review. Future research can explore adapting the methodology to different contexts, expanding the range of fuzzy criteria, and integrating additional sources of human feedback to further enhance alignment and robustness.

\section{Conclusion and Future Work}
This study introduces LLM-as-a-Fuzzy-Judge, a hybrid evaluation framework that combines SFT and prompt engineering to align LLM judgments with human expert preferences in medical education. Our results demonstrate that the hybrid approach achieves strong agreement with human judges on multiple subjective fuzzy criteria with high accuracy in particularly at major levels, including professionalism, medical relevance, ethical behavior, and contextual distraction. The model consistently outperforms prompt-only and majority-class baselines, providing nuanced, interpretable, and human-aligned feedback in complex clinical scenarios. The LLM-as-a-Fuzzy-Judge offers an additional solution to save the teacher's time evaluating students' skills. 

Future work would focus on expanding the diversity and scale of annotated datasets to further improve the robustness and generalizability of the model. Incorporating additional fuzzy criteria or domain-specific dimensions could enhance the system's applicability to broader educational and professional contexts. Exploring advanced alignment techniques, such as RLHF, active learning and fuzzy interference, may further refine the model's ability to capture subtle human preferences. Finally, integrating the LLM-as-a-Fuzzy-Judge into real-world educational platforms (2-Sigma) and conducting longitudinal studies will be essential to assess its impact on learner outcomes and faculty assessment workflows.
%%%%%%%%%%%%%%%%%%%%%%%%%%%%%%%%%%%%%%%%%%%%%%%%

\begin{credits}
\subsubsection{\ackname} This study was funded by the American Medical Association (AMA grant) for 2-Sigma Research and technical support from the Advanced Research Computing Center (ARC) at the University of Cincinnati. The manuscript and code used generative AI to polish and help with the writing, coding, and training. We thank all medical experts for annotating the data and reviewers for reviewing the article.

%\subsubsection{\discintname} The authors have no competing interests.

% \subsubsection{\discintname}

\end{credits}
\bibliographystyle{splncs04}
\bibliography{ref}

\end{document}